\documentclass[sigconf]{cidr-2026}
\usepackage{hyperref}
\usepackage{hyperxmp}
\usepackage{cleveref}
\begin{document}

\title{Deep Research is the New Analytics System:\\Towards Building the Runtime for AI-Driven Analytics}


% \author{Matthew Russo, 
% {Vikram Nathan\textsuperscript{\textdagger}, Murali Narayanaswamy\textsuperscript{\textdagger}, Tim Kraska}\\ 
% 	\normalsize {MIT, \textsuperscript{\textdagger} Amazon\\}
%  \texttt{mdrusso@csail.mit.edu, vrnathan,muralibn@amazon.com, kraska@mit.edu}\\
%  }
\author{Matthew Russo}
\orcid{0009-0005-9685-3976}
\affiliation{%
  \institution{MIT}
  \city{Cambridge}
  \state{MA}
  \country{USA}
}
\email{mdrusso@csail.mit.edu}

\author{Tim Kraska}
\affiliation{%
  \institution{MIT}
  \city{Cambridge}
  \state{MA}
  \country{USA}
}
\email{kraska@mit.edu}

% \author{Murali Narayanaswamy}
% \affiliation{%
%   \institution{Amazon}
%   \city{San Jose}
%   \state{CA}
% \email{muralibn@amazon.com}

% \author{Vikram Nathan}
% \affiliation{%
%   \institution{Amazon}
%   \city{San Jose}
%   \state{CA}
% \email{vrnathan@amazon.com}

\renewcommand{\shortauthors}{Russo et al.}
\newcommand{\matt}[1]{\textcolor{blue}{#1}}
\newcommand{\tim}[1]{\textcolor{red}{#1}}

\begin{abstract}
With advances in large language models (LLMs), researchers are creating new systems that can perform AI-driven analytics over large unstructured datasets. Recent work has explored executing such analytics queries using semantic operators---a declarative set of AI-powered data transformations with natural language specifications. However, even when optimized, these operators can be expensive to execute on millions of records and their iterator execution semantics make them ill-suited for interactive data analytics tasks. In another line of work, Deep Research systems have demonstrated an ability to answer natural language question(s) over large datasets. These systems use one or more LLM agent(s) to plan their execution, process the dataset(s), and iteratively refine their answer. However, these systems do not explicitly optimize their query plans which can lead to poor plan execution. In order for AI-driven analytics to excel, we need a runtime which combines the optimized execution of semantic operators with the flexibility and more dynamic execution of Deep Research systems. As a first step towards this vision, we build a prototype which enables Deep Research agents to write and execute optimized semantic operator programs. We evaluate our prototype and demonstrate that it can outperform a handcrafted semantic operator program and open Deep Research systems on two basic queries. Compared to a standard open Deep Research agent, our prototype achieves up to 1.95x better F1-score. Furthermore, even if we give the agent access to semantic operators as tools, our prototype still achieves cost and runtime savings of up to 76.8\% and 72.7\% thanks to its optimized execution.
\end{abstract}
% Inspired by relational operators and query optimization, semantic operator systems can optimize these queries to run more efficiently.

\maketitle

\begin{figure*}
\includegraphics[width=\textwidth]{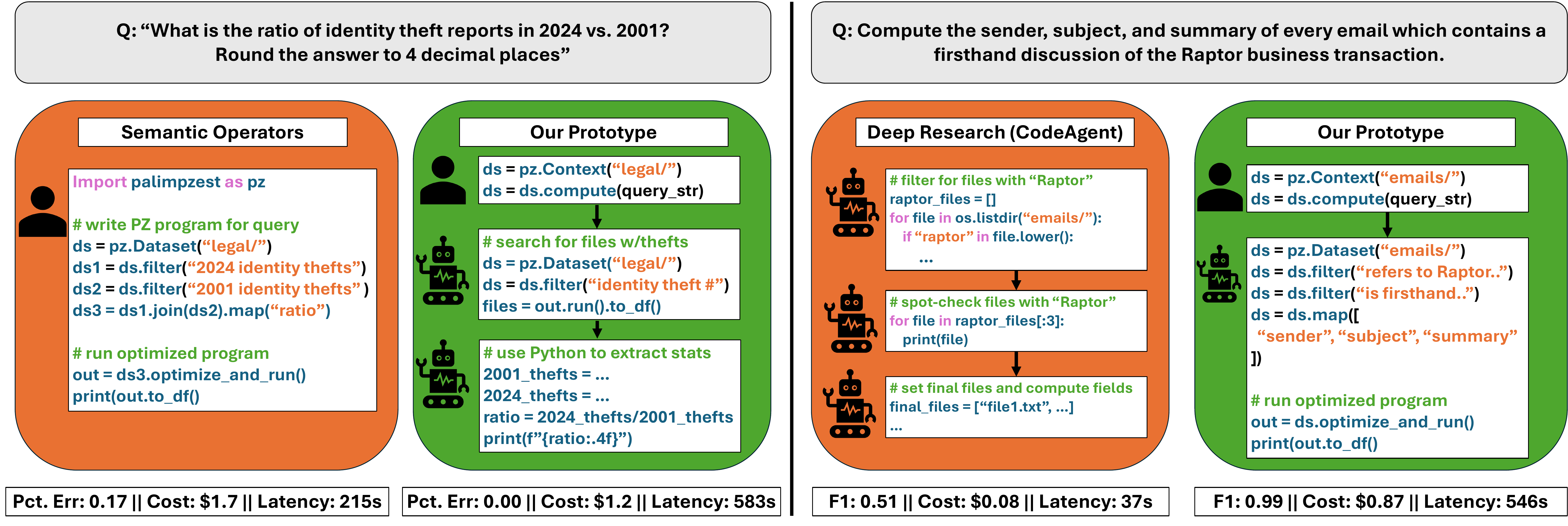}
\caption{(Left) An example query from the Kramabench dataset which a handcrafted semantic operator program struggles to perform well on. Our prototype iterates between executing optimized semantic operator programs and writing Python code to identify the correct statistics and compute the final ratio. (Right) An example query on the Enron email dataset. An open Deep Research system achieves low recall (and F1-score) because it filters the data using simplistic Python code and only returns the few emails which it manually validates. Our prototype writes an optimized semantic operator program to process the entire dataset thus improving recall and F1-score.}
\Description{(Left) An example query from the Kramabench dataset which a handcrafted semantic operator program struggles to perform well on. Our prototype iteratively writes Python code to identify the correct statistics and compute the final ratio. (Right) An example query on the Enron email dataset. An open Deep Research system achieves low recall (and F1-score) because it filters the data using simplistic Python code and only returns the few emails which it manually validates. Our prototype writes an optimized PZ program to process the entire dataset thus improving recall and F1-score.}
\label{fig:example}
\end{figure*}

\section{Introduction}
Enabling (complex) analytics over large unstructured data lakes has been a long-standing goal of data systems research. Traditional OLAP databases \cite{cstore, redshift, snowflake, duckdb}, while great for structured queries, have struggled to support these workloads. 
SQL as the primary language is not sufficient to interact with the vast variety of unstructured data. 
To address this limitation, researchers --- mostly from the database community -- have proposed semantic operators—AI-powered analogs to relational operators inspired by classical query optimization techniques. Systems like Palimpzest \cite{liu2025palimpzest,russo2025abacuscostbasedoptimizersemantic}, LOTUS \cite{patel2025semanticoperatorsdeclarativemodel}, DocETL \cite{shankar2024docetlagenticqueryrewriting}, Galois \cite{galois2025satriani}, and others \cite{lu2025vectraflow, aryn2025, urban2023caesuralanguagemodelsmultimodal} enable developers to apply operations such as AI-driven maps, filters, joins, and aggregations—specified in natural language—over large unstructured datasets. Early research shows that these semantic operators can be effectively optimized for a variety of tasks, including information extraction, summarization, ranking, and more \cite{russo2025abacuscostbasedoptimizersemantic}.

In contrast, the AI-community went down a different route: Deep Research systems \cite{claude-research, gemini-deep-research, openai-deep-research, perplexity-deep-research, langchain-open-deep-research, smolagents-open-deep-research, xia-deep-research}, which are able to create (Python) code on the fly to query structured and unstructured data alike. Open Deep Research systems, like HuggingFace's SmolAgents implementation \cite{smolagents-open-deep-research}, use so-called ``CodeAgents" \cite{smolagents-code-agent} that can reason, write code, and use tools in an iterative fashion to execute a natural language instruction. 
%%%However, even with optimization, they can be expensive to execute on datasets with millions of records, and their iterator execution semantics make them ill-suited for data analytics tasks requiring dynamic interaction with data.

%are woefully insufficient for many Deep Research queries.
% Early work in this direction focused on extending SQL to support unstructured analytics with customized operators for tasks such as image classification, object detection, sentiment analysis, and more \cite{Anderson2018PhysicalRP, kang2017noscope, kang2019blazeit, russo2023inquest, kang2021abae}.

%%% The interactive / dynamic execution semantics of these agents help them succeed on data analytics questions which cannot easily be answered semantic operator programs.

When compared at a high-level,  OLAP systems, semantic operator systems, and Deep Research systems are all remarkably similar. Given a user's declarative query, each system creates a query plan, executes it over a dataset, and returns the final result. All systems have access to many query plans, but ideally prefer ones which are fast, cheap, and accurate. Finally, in each setting, predicting plan performance is non-trivial, and optimizing the query plan is critical for good performance. However, Deep Research systems are in many ways much more flexible; not only do they take natural language as an input but also often used techniques like self-reflection and incremental planning, which are best compared to adaptive query processing on steroids.  

\textbf{Challenges.} However, semantic operator systems and Deep Research systems each have their own weaknesses which make neither framework fully adequate for answering unstructured analytics queries. For example, the left-hand side of \Cref{fig:example} shows a query from the recently published Kramabench \cite{lai2025kramabenchbenchmarkaisystems} benchmark. The dataset consists of 132 files with statistics on fraud, identity theft, and other consumer reports, and the query asks the analytics system to compute the ratio of identity theft reports over a two year span. A hand-crafted semantic operator program tries to answer the query by filtering for files with statistics on identity theft reports, before using a map to compute the ratio.

% TODO: rework
The main issue with this program is that semantic operators' execution semantics can only process one file at a time. In addition to being expensive and slow (the majority of files in this dataset have state-level statistics which can be ignored for this query) this makes writing a correct program for this query difficult. A handful of files contain 2024 statistics and inferring which file is correct for this query requires reasoning across multiple files simultaneously. As a result, this program is expensive, slow, and often produces incorrect output(s).

While Deep Research systems can process the previous query with relative ease, they also struggle on other queries which are simple for semantic operators. For example, the right-hand side of \Cref{fig:example} shows a query from the Enron email dataset \cite{enron}. This query asks the analytics system to return all emails with firsthand discussion of certain business transactions. The Deep Research system achieves high precision and low recall by writing Python code to search for certain keywords and then manually reading and verifying a few of those emails. By comparison, a handcrafted semantic operator program can answer this query almost perfectly. % (implemented with a SmolAgents' CodeAgent \cite{smolagents-code-agent})

While the Deep Research system's high-level query plan---filter for potentially relevant emails and verify that they satisfy the query---is not unreasonable, its execution of the query plan is poor. Especially when processing large datasets, we observe that Deep Research agents have a tendency to ``take shortcuts" and do not consider expensive strategies such as reading every file to apply the filter predicate. Another common failure mode involves the agent terminating prematurely before it has fully executed its plan, which has also been observed elsewhere \cite{cemri2025multiagentllmsystemsfail}.

\textbf{Goal.} Hence we argue, that an ideal runtime for AI-driven analytics would utilize the strengths of Deep Research, semantic operators, and OLAP analytics systems. This runtime would share Deep Research's flexibility to write an initial query plan, iteratively execute Python code and use tools, and dynamically update its query plan in response to observations. It would also share semantic operator systems' ability to optimize their query plans automatically, using cost-based optimization \cite{russo2025abacuscostbasedoptimizersemantic}, query rewrites \cite{shankar2024docetlagenticqueryrewriting}, and physical operator optimizations \cite{patel2025semanticoperatorsdeclarativemodel}. Finally, this runtime should leverage structured information, possibly generated from unstructured data, which it can then query using SQL. This functionality is especially important in analytics settings where many queries are issued against the same data lake and future queries can reuse structured tables which are generated to answer previous queries.

% The key difference between Deep Research and OLAP database systems is the guarantees (or lack thereof) that each system provides with respect to accuracy. Leaving approximate query processing (AQP) \cite{pilotdb, kang2021abae, kang2019blazeit, russo2023inquest, Anderson2018PhysicalRP, kang2017noscope} aside for the moment, standard OLAP databases perform deterministic computation over structured columns and tables. On the other hand, Deep Research systems use artificial intelligence in some capacity to analyze unstructured inputs, write code, and query data sources. The introduction of AI removes most (if not all) guarantees about correctness, given its propensity for reasoning mistakes and hallucinations.

% In spite of its lack of correctness guarantees, Deep Research provides value to users with its simplicity: upload your documents, connect your data sources, and ask a question. There is no need for users to define tables, schemas, primary keys, indices, etc. They simply provide their data to the system, and let its query engine figure out the best way to compute the answer. 

% In a span of five months, demand for these systems led OpenAI, Anthropic, Google, Perplexity, and xAI \cite{openai-deep-research, claude-research, gemini-deep-research, perplexity-deep-research, xia-deep-research} to roll out offerings to their combined hundreds of millions of users. Given the size and importance of OLAP systems within the Databases industry, one can imagine a future where Deep Research grows to become a similarly prominent pillar of the broader AI industry.

\textbf{Our Approach.} As a first step towards building a unified runtime for AI-driven analytics, we created a prototype which aims to combine the flexibility of Deep Research with the optimized execution of semantic operators and SQL. Our prototype extends the Palimpzest framework \cite{liu2025palimpzest} in three key ways. First, we introduced two new semantic operators: \texttt{compute} and \texttt{search}, which are physically implemented with agents that can plan their execution, write code, and use tools. Crucially, each operator is provided with a tool that can execute a natural language instruction with an optimized semantic operator program. This imbues each agent with the ability to execute optimized query plans as part of its more dynamic planning and execution.
% \textbf{Our Approach.} As a first step towards building a unified runtime for AI-driven analytics, we created a prototype which aims to combine the flexibility of Deep Research with the optimized execution of semantic operators and SQL. To accomplish this, we made three key additions to the Palimpzest framework~\cite{liu2025palimpzest}. First, we introduced Deep Research agents to the framework by adding two new semantic operators: \texttt{compute} and \texttt{search}. Each operator takes an input dataset and a natural language description of the data to compute, or search for, respectively. Each operator is then compiled to a CodeAgent by Palimpzest's query optimizer. The \texttt{compute} operator's CodeAgent is also provided with a tool which can take a natural language instruction, write an optimized Palimpzest program, and execute it. This enables the agent to execute an optimized semantic operator program as a step within its larger execution trace when the agent deems necessary for accomplishing the \texttt{compute} operator's larger instruction.

For example, on the right-hand side of \Cref{fig:example}, our prototype implements the Enron email query by passing the query string to a \texttt{compute} operator. The operator performs a few data exploration steps (not shown) before invoking its tool for writing a semantic operator program. That program ultimately executes with near-perfect accuracy, and---as shown in our evaluation in \Cref{sec:evaluation}---is more efficient than providing an open Deep Research system with unoptimized semantic operators as tools.

Second, we extended Palimpzest's \texttt{Dataset} abstraction, to support access methods beyond iteration over a set of records. Specifically, we added support for users to provide their own indexing and top-k methods on top of their custom datasets. This more general abstraction for data access, which we call a \texttt{Context}, also contains a natural language description of the data it encompasses. Finally, users may also add their own tools to the \texttt{Context} to enable agents to interact with data in more bespoke ways. Such tools can include additional data access methods (e.g., a web search tool or a secondary index over a data lake), as well as methods for data cleaning, data visualization, etc.

Third, inspired by the use of materialized views in OLAP databases, we introduce a mechanism for indexing \texttt{Contexts} that are materialized during query execution. Each time a \texttt{compute} or \texttt{search} operator executes an instruction it generates a new \texttt{Context}, which is akin to a materialized view over the original \texttt{Context}. In an effort to leverage past query execution(s) to optimize new queries, we enabled Palimpzest's query optimizer to retrieve previously materialized \texttt{Contexts} which achieve high similarity to new \texttt{compute} and \texttt{search} instructions. % we also added an optimization for \texttt{compute} and \texttt{search} operators which retre additional relevant \texttt{Contexts} which the operator may use.

\textbf{Contributions.} In this paper we present a vision for building a new runtime for AI-driven analytics that combines the efficiency of SQL and semantic operators with the power and flexibility of Deep Research systems. In particular, we:

\begin{itemize}
    \item Extend Palimpzest to include \texttt{search} and \texttt{compute} operators which support iterative execution patterns commonly found in Deep Research. (\Cref{sec:overview}.)
    \item Introduce a new \texttt{Context} abstraction, which enables more dynamic access patterns in Palimpzest. (\Cref{sec:overview}.)
    \item Describe physical and logical optimizations for \texttt{search} and \texttt{compute} operators. (\Cref{sec:optimization}.)
    \item Demonstrate that our prototype can execute an evaluation query that semantic operators struggle to process. (\Cref{sec:evaluation}).
    \item Demonstrate that our prototype achieves 1.95x better F1-score than an open Deep Research system on a query from Kramabench, and achieves cost and runtime savings of 76.8\% and 72.7\%, respectively, relative to a Deep Research system which uses semantic operators as tools. (\Cref{sec:evaluation}).
\end{itemize}

% \begin{itemize}
%     \item However, as illustrated in Figure 1 (left column), Deep Research can still struggle to answer fairly simple questions, and--even when provided with semantic operators as tools (middle column)--it struggles to use them efficiently
%     \item \textbf{Goal.} We want to build a runtime for AI-driven analytics which has best of both worlds: the efficient, optimized execution of SQL and semantic operators, with the flexibility and dynamism of Deep Research. % see language in Notes
%     \item \textbf{Our Approach.} To accomplish this, we explored giving a CodeAgent the ability to write optimized PZ programs. As illustrated in Figure 1 (right column) CodeAgent is able to write a program which is optimized by PZ to efficiently achieve good performance.
%     \item Our implementation, which we built inside of the Palimpzest library, consists of three main parts... (context, compute/search, context manager). 
%     \item \textbf{Contributions.} We set a vision for building a new runtime for AI-driven analytics that combines the efficiency of SQL / semantic operators with the power and flexibility of Deep Research. We build an initial prototype in PZ and demonstrate 
% \end{itemize}
\begin{figure*}
\includegraphics[width=\textwidth]{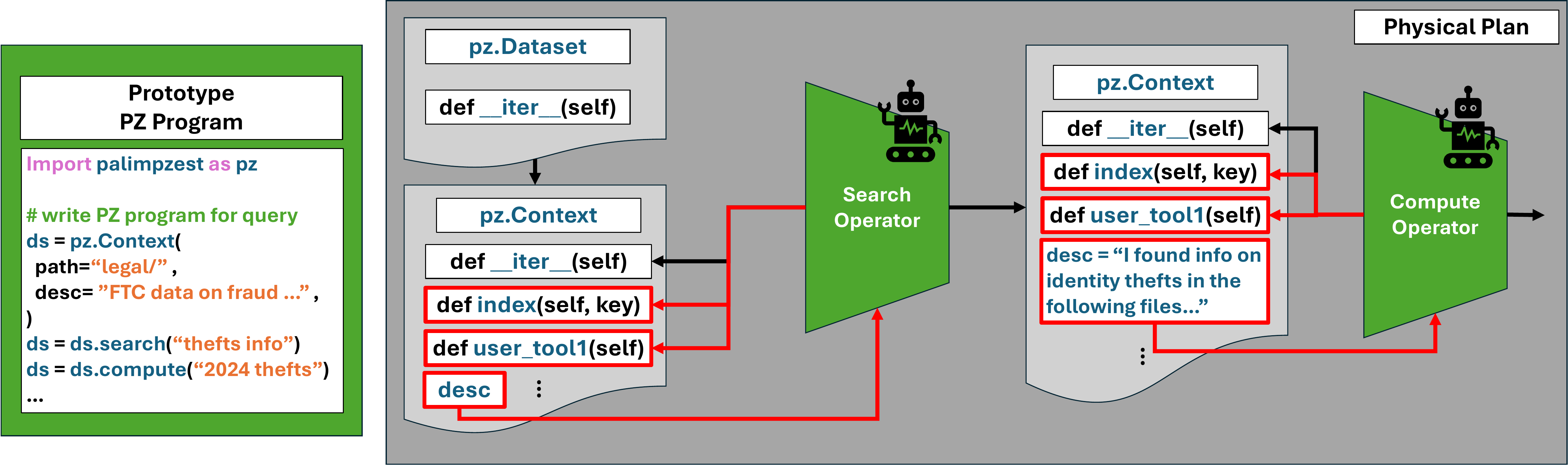}
\caption{Overview of a Palimpzest (PZ) program and its physical plan. The program creates an initial \texttt{Context} object, which the \texttt{search} operator uses to look for information on identity thefts. The \texttt{search} operator outputs a new \texttt{Context} object with an updated description reflecting the results of its search. Finally, the \texttt{compute} operator uses this intermediate \texttt{Context} to compute the number of identity thefts in 2024.}
\Description{Overview of a Palimpzest (PZ) program and its physical plan. The program creates an initial \texttt{Context} object, which the \texttt{search} operator uses to look for information on identity thefts. The \texttt{search} operator outputs a new \texttt{Context} object with an updated description reflecting the results of its search. Finally, the \texttt{compute} operator uses this intermediate \texttt{Context} to compute the number of identity thefts in 2024.}
\label{fig:overview}
\end{figure*}

\section{Overview}
\label{sec:overview}
We present an overview of our prototype. First, we provide a quick background on implementing analytical queries with semantic operators and Deep Research systems. Then we describe our new \texttt{Context} abstraction as well as the \texttt{search} and \texttt{compute} operators which enable agents to execute optimized semantic operator programs. Finally, we discuss the addition of a \texttt{ContextManager} which indexes materialized \texttt{Contexts} so that they may be used to optimize future queries.

\subsection{Analytical Queries with Semantic Operators and Deep Research systems}
Recent work has focused on using semantic operators to execute analytical queries over unstructured datasets \cite{liu2025palimpzest, patel2025semanticoperatorsdeclarativemodel, shankar2024docetlagenticqueryrewriting, aryn2025, galois2025satriani, lu2025vectraflow, urban2023caesuralanguagemodelsmultimodal}. Unlike relational operators which are specified with SQL expressions, these operators are specified in natural language and are useful for a range of document processing tasks \cite{russo2025abacuscostbasedoptimizersemantic, liu2025palimpzest, patel2025semanticoperatorsdeclarativemodel}. However, similar to their relational counterparts, semantic operators have iterator execution semantics which make them inefficient for many queries. For example, in the semantic operator program on the left-hand side of \Cref{fig:example}, the first semantic filter will process all 132 files in the dataset even after it has found the number of identity thefts in the year 2024. 

By contrast, Deep Research systems are capable of writing execution plans which enable them to more efficiently access the data they need. For example, on the same query a SmolAgents CodeAgent will list all 132 files and use information in the filename to determine which file(s) to read. However, while Deep Research agents are adept at writing query plans, their execution of these plans is often suboptimal. For example, an agent may generate a plan to read every file until it finds the file with identity thefts in 2024, and then give up on reading the dataset after the fourth or fifth file. These observations led us to consider whether we could combine the optimized execution of semantic operators with the planning and dynamic execution of Deep Research agents.

% First, writing a query with semantic operators requires the user to be aware of the data sources in their data lake and possibly, as in \Cref{fig:example}, to know \textit{a priori} how data is laid out across different files. Many Deep Research queries require the system to perform data discovery on its own, which can only be implemented... Finally, even if a Deep Research query can be expressed with a semantic operator program, its record-at-a-time execution can be highly inefficient for queries which only pertain to a small number of data items in the data lake.

% These challenges motivate a new set of abstractions for defining computation over unstructured data lakes.

\subsection{Context Class}
In order to support more efficient query execution patterns, we created a new \texttt{Context} abstraction in Palimpzest. This abstraction, illustrated in \Cref{fig:overview}, inherits from Palimpzest's \texttt{Dataset} class which means it automatically supports iterator execution. Additionally, the \texttt{Context} class also exposes an index method which allows programmers to define key-based point lookups and/or vector-based search on their own dataset(s). Programmers can also define custom tools which are relevant for their dataset. For example, a programmer working with time-series data might expose tools for accessing sequence data at different sample frequencies. Finally, the \texttt{Context} class also has a description field (``desc"), which contains a natural language description of the dataset's contents.

While supporting various access methods, tools, and a description may seem arbitrary, these features are essential for enabling agents to interact with the dataset. By exposing iterator and index-based access methods, agents can make informed decisions about which access pattern to use when executing a given query. In the example from before, an agent may decide to use an embedding-based lookup to search for files with the number of identity thefts in 2024. Then, based on the result of its search, it can either use the information that was returned or resort to iterating over the entire dataset. Furthermore, detailed descriptions can assist agents in deciding which file(s) within a \texttt{Context} are likely to contain the information they need.

\subsection{Compute and Search Operators}
While the \texttt{Context} abstraction may be useful for agents, leveraging this abstraction requires building an operator which actually uses an agent. To address this, we set out to create a logical operator which could be compiled by Palimpzest and physically implemented with an agent. Ultimately, our experience trying to execute various queries from Kramabench \cite{lai2025kramabenchbenchmarkaisystems} led us to create two logical operators: one for \texttt{compute} and one for \texttt{search}. Each operator takes a \texttt{Context} as input, but they have slightly different semantics: the \texttt{compute} operator seeks to generate a specific output, whereas the \texttt{search} operator tries to find information which can be used to enrich a \texttt{Context}'s description. Finally, given the widespread adoption of SmolAgent's CodeAgent, we decided to have a CodeAgent to serve as our physical implementation for each logical operator.

We illustrate the functionality of these new operators in \Cref{fig:overview}. The program on the left-hand side of the diagram creates an initial \texttt{Context} object, which includes a natural language description of the data along with support for indexing and tool use. The \texttt{search} operator takes the \texttt{Context} as input, and uses its description, tools, and data access methods to search for information on identity thefts. The operator outputs a new \texttt{Context} and updates its description to contain (a summary of) its search execution trace. This \texttt{Context} is then fed into a \texttt{compute} operator, which reads the description and uses the tools and access methods to compute the number of thefts in 2024.

\subsection{Context Management \& Maintenance}
One of the key challenges with query processing using LLMs is the high cost and latency associated with invoking the models. For instance, processing 1,000 emails to extract their sender and subject is orders of magnitude more expensive and time consuming than executing a SQL query over a structured table with sender and subject columns. Prior work on optimizing LLM inference has shown that a simple way reduce LLM computation is to reuse previous execution results \cite{zheng2024sglangefficientexecutionstructured}. Similarly, databases have long used materialized views to reduce unnecessary computation \cite{views}. Thinking along these lines, we created a \texttt{ContextManager} which embeds and caches the descriptions of materialized \texttt{Contexts}. This enables Palimpzest's optimizer to reuse these \texttt{Contexts} to help with answering future queries, which we discuss further in the next section.
\section{Optimization}
\label{sec:optimization}
The abstractions presented in the previous section provide systems like Palimpzest with a number of opportunities for (cost-based) optimization. In this section, we describe an initial set of logical and physical optimization which we have already implemented in our prototype, or plan to implement in the near future.

\textbf{Logical Optimizations.} Inspired by work from DocETL \cite{shankar2024docetlagenticqueryrewriting}, an obvious avenue for logical optimization is to rewrite the query plan to better scope \texttt{search} and \texttt{compute} directives which are underspecified or overly complex. DocETL has demonstrated success at rewriting data processing pipelines in this fashion, and we believe Palimpzest's query optimizer could similarly apply an LLM judge to determine when a \texttt{search} or \texttt{compute} operator needs to be split into smaller operations. Alternatively, the query optimizer may also be able to improve the efficiency of the query plan by identifying opportunities to merge \texttt{search} and \texttt{compute} operations which are similar, or likely to process the same set of physical data. Finally, for \texttt{compute} operations which repeatedly fail during query execution, a dynamic query optimizer could insert a logical \texttt{search} operator at runtime before the \texttt{compute} in an attempt to provide it with a richer \texttt{Context}. All of the aforementioned optimizations are currently future work, and outside the scope of our prototype.

\textbf{Physical Optimizations.} Similar to work in Abacus \cite{russo2025abacuscostbasedoptimizersemantic}, one direction for physical optimization includes allowing the query optimizer to select the model (and other parameters) used by the \texttt{compute} or \texttt{search} operator's agent. Building on ideas from materialized views research \cite{views}, another opportunity for optimization is to reuse previously cached \texttt{Context(s)} to augment (or replace) the \texttt{Context} for a \texttt{compute} operator. For example, if a query computes the identity theft reports in 2001, and then a second query seeks to compute the identity thefts in 2024, a detailed \texttt{Context} from the first query might help the second query execute more efficiently. We have implemented a preliminary version of this physical optimization in our prototype, although it is currently experimental.

\section{Evaluation}
\label{sec:evaluation}
We evaluate our prototype system on two example queries. The first query comes from the \texttt{legal} workload in the Kramabench \cite{lai2025kramabenchbenchmarkaisystems} benchmark. The workload's dataset consists of 132 CSV and HTML files which contain recent statistics on fraud, identity theft, and other consumer reports. The query asks the analytics system to compute the ratio between the number of identity theft reports in the years 2024 and 2001. The ground truth for this query is found in a single CSV file, which contains a breakdown of the fraud, identity theft, and other reports for the years 2001 to 2024 inclusive.

The second query asks the analytics system to filter a subset of emails from the Enron email dataset \cite{enron} for ones which contain firsthand discussion of one or more specific business transactions. Similar to prior work \cite{liu2025palimpzest}, we evaluate on a subset of 250 Enron emails in order to keep execution costs reasonable. The query also asks the system to extract a sender, subject, and summary of each email, but to simplify our evaluation we simply compute the precision, recall, and F1-score of the emails returned by each system.

\textbf{Our Prototype Outperforms Semantic Operators.} For our first experimental claim, we sought to demonstrate that the dynamic execution semantics of our prototype's \texttt{compute} operator could outperform a handcrafted semantic operator program on an interactive data analytics tasks. For this evaluation, we use the \texttt{legal-easy-3} query from the Kramabench \cite{lai2025kramabenchbenchmarkaisystems} dataset. For our baselines, we handcrafted a Palimpzest program and a SmolAgents CodeAgent equipped with tools for listing and reading files. For our prototype, we execute the \texttt{compute} operator by simply passing in the query string as the natural language instruction.

\begin{table}
    \centering
    \caption{\texttt{compute} achieves lower error than a handcrafted semantic operator program written in Palimpzest. In trials where the semantic operator program returned multiple output ratios, we averaged the percent errors for each ratio.}
    % \vspace{-12pt}
    \renewcommand{\arraystretch}{1.0}
    \begin{tabular}{|l|r|r|r|r|r|}
    \hline
    {\bf System} & {\bf Pct. Err.} & {\bf Cost (\$)} & {\bf Time (s)} \\
    \hline
    Sem. Ops & 17.00\% & 1.66 & 215.2 \\
    \hline   
    CodeAgent & 27.56\% & 0.03 & 77.0 \\
    \hline   
    PZ \texttt{compute} & {\bf 0.02\%} & 1.17 & 583.0 \\
    \hline   
    \end{tabular}
    \label{tab:krama-results}
    % \vspace{-18pt}
\end{table}
We ran each system on the evaluation query three times and report the average percent error, cost, and runtime. The results from our evaluation are shown in \Cref{tab:krama-results}.

Overall, our prototype's \texttt{compute} operator achieved the lowest percent error across all three systems. The semantic operator program was able to compute the correct ratio in all three trials, however in two trials it also computed a second ratio due to an errant file returned by one of its semantic filters. The CodeAgent often struggled to find the correct file in the dataset, and would return spurious ratios which it computed from the information it found in non-ground truth files. Finally, our prototype was able to effectively answer the query by writing optimized PZ programs to search for information on identity thefts in 2024 and 2001, before computing the final result in Python.

\textbf{Our Prototype Outperforms Open Deep Research.} For our second experimental claim, we sought to demonstrate that our prototype's \texttt{compute} operator could leverage its ability to write optimized PZ programs to outperform a SmolAgents CodeAgent. For the evaluation, we recreated a document processing task from \cite{liu2025palimpzest} in which each system has to filter for emails matching two natural language predicates. For our baselines, we use two CodeAgents: the first CodeAgent is equipped with tools for listing and reading files, while the second agent (CodeAgent+) is additionally equipped with tools for applying (unoptimized) semantic filters and maps. For our prototype, we execute the \texttt{compute} operator by passing in the query string as the natural language instruction.

We ran each system on the evaluation query three times and report the average quality, cost, and runtime. We implemented each system with GPT-4o and used GPT-4o in CodeAgent+'s semantic operator tools as well. The results from our evaluation are shown in \Cref{tab:enron-results}.
\begin{table}
    \centering
    \caption{\texttt{compute} writes optimized Palimpzest (PZ) programs which achieve higher quality than a naive CodeAgent and have better optimized execution than a CodeAgent with semantic operators as tools (CodeAgent+).}
    % \vspace{-12pt}
    \renewcommand{\arraystretch}{1.0}
    \begin{tabular}{|l|r|r|r|r|r|}
    \hline
    {\bf System} & {\bf F1} & {\bf Recall} & {\bf Prec.} & {\bf Cost (\$)} & {\bf Time (s)} \\
    \hline        
    CodeAgent & 50.53\% & 46.15\% & 88.89\% & 0.08 & 37.0 \\
    \hline   
    CodeAgent+ & 98.67\% & 97.44\% & 100\% & 3.76 & 1,999.9 \\
    \hline   
    PZ \texttt{compute} & {\bf 98.67\%} & 97.44\% & 100\% & 0.87 & 546.2 \\
    \hline   
    \end{tabular}
    \label{tab:enron-results}
    % \vspace{-18pt}
\end{table}
Overall, our prototype's \texttt{compute} operator achieved near perfect output quality while saving 76.8\% on cost and 72.7\% on runtime relative to CodeAgent+ which manually invokes semantic operators. We observe that the CodeAgent without semantic operators is cheap and fast, but fails to achieve good quality (i.e. F1-score) on the objective. The agent’s low quality---driven by poor recall of relevant emails---is a result of its tendency to rely on simple filter heuristics (e.g., regular expressions) and manual reading to identify emails that are relevant.

Providing CodeAgent+ with semantic operators as tools helps the agent overcome the aforementioned recall issue. By invoking semantic map and filter operations, CodeAgent+ guarantees that every email is read and processed by an LLM. This leads to a significant boost in quality, however it also increases the cost and runtime of the system. Unfortunately, much of the increase in the cost and runtime is due to inefficient use of semantic operators. For instance, CodeAgent+ often executed multiple semantic filters in sequence without checking the output of the first semantic filter before executing the subsequent one(s).

Finally, our \texttt{compute} operator wrote correct Palimpzest programs in all three trials. These programs were immediately more efficient than those written by CodeAgent+ because Palimpzest's execution engine did not perform redundant computation on filtered emails. Furthermore, Palimpzest's query optimizer was able to use cheaper models for some of the semantic operators, thus yielding additional cost and runtime savings.

% We implement the 
% \begin{itemize}
%     \item Run all 30 queries from Kramabench's legal dataset with PZ (semantic operators only) and PZ + \texttt{compute} and \texttt{search}
%     \begin{itemize}
%         \item \textbf{Note: I have already done the latter; I would just need to write and execute the queries with semantic operators}
%         \item Ideally, we show that \texttt{compute} and \texttt{search} perform much better in terms of quality, cost, and runtime
%     \end{itemize}
%     \item Run all 30 queries from Kramabench's legal dataset (PZ + \texttt{compute} and \texttt{search}) in sequence, with and without the ContextManager
%     \begin{itemize}
%         \item \textbf{Note: I have also run this experiment, although the \texttt{search} operator needs to be improved for this to work.}
%         \item Ideally, we show that executing the queries in sequence with the ContextManager leads to cost and runtime savings because the \texttt{search} operator can benefit from previously computed Contexts which describe where certain data lives.
%     \end{itemize}
% \end{itemize}
\section{Related Work}
Deep Research Systems---which answer natural language questions over large datasets using a combination of LLM planning, tool use, code execution, and reasoning---have become increasingly popular for data analytics tasks. In a span of five months, OpenAI, Anthropic, Google, Perplexity, and xAI \cite{openai-deep-research, claude-research, gemini-deep-research, perplexity-deep-research, xia-deep-research} all released private offerings to their combined hundreds of millions of users. Since then, researchers in industry and academia have sought to replicate the performance of these systems.

Some of the more prominent open source offerings include LangChain's Open Deep Research \cite{langchain-open-deep-research} and HuggingFace's SmolAgents Open Deep Research \cite{smolagents-open-deep-research}. LangChain's system uses LLMs (agents), search tools, and MCP server(s) in three modular phases to scope, research, and write a final report for the user's question. HuggingFace's system uses its CodeAgents \cite{smolagents-code-agent} to iteratively plan, use tools, and execute Python code in order to answer the user's question. Our prototype's \texttt{compute} and \texttt{search} operators are physically implemented with CodeAgents which are provided with a tool for writing optimized Palimpzest programs.

Another recent line of work focuses on optimizing and executing analytics queries over unstructured data with semantic operator systems \cite{liu2025palimpzest, patel2025semanticoperatorsdeclarativemodel, shankar2024docetlagenticqueryrewriting, aryn2025, lu2025vectraflow, urban2023caesuralanguagemodelsmultimodal, galois2025satriani}. Palimpzest \cite{liu2025palimpzest, russo2025abacuscostbasedoptimizersemantic} uses a multi-armed bandit sampling algorithm to gather statistics on operator performance before using cost-based optimization to optimize the semantic operator system. LOTUS \cite{patel2025semanticoperatorsdeclarativemodel} uses proxy methods to optimize semantic join, filter, group-by, and top-k operators while providing statistical guarantees with respect to a reference model. DocETL \cite{shankar2024docetlagenticqueryrewriting} uses LLMs to apply (and validate) query rewrites to semantic operator systems. Finally, Galois \cite{galois2025satriani} introduced new logical and physical optimizations for answering queries with LLM-based operators. Our work extends these systems to support Deep Research queries requiring more dynamic execution semantics.

While traditional OLAP databases \cite{cstore, redshift, snowflake, duckdb} have limited support for AI-driven analytics over unstructured data, there is considerable work on Approximate Query Processing (AQP) which aims to support these queries. Some early work focused on adding machine learning classifiers to analytics systems for tasks including object detection, sentiment analysis, and more \cite{Anderson2018PhysicalRP, kang2017noscope, kang2019blazeit, russo2023inquest, kang2021abae}.  One limitation of these systems was they struggled to answer queries which did not align well with the task the classifier was trained for. While our prototype and AQP share a similar tradeoff between efficiency and execution accuracy, the generality of foundation models and dynamic execution semantics of \texttt{compute} and \texttt{search} enable our system to answer a broader range of analytics queries.

\section{Conclusion}
We present a vision and a prototype for a new runtime for AI-driven analytics over large unstructured datasets. In particular, we propose a runtime which blends the optimized execution of semantic operator systems with the flexibility and dynamic execution semantics of Deep Research systems. To this end, we extend the Palimpzest framework to include two new semantic operators which leverage CodeAgents to execute a natural language instruction. To support these operators, we create a new \texttt{Context} abstraction in Palimpzest which enables dynamic data access methods such as indexing and search. Finally, we demonstrate our prototype's ability to leverage the strengths of Deep Research and optimized semantic operator execution on two evaluation queries for interactive data analytics and document processing, respectively.

\bibliographystyle{ACM-Reference-Format}
\bibliography{sample-base}

% %%
% %% If your work has an appendix, this is the place to put it.
% \appendix









\end{document}